\documentclass[10pt,twocolumn,letterpaper]{article}

\usepackage{cvpr}
\usepackage{times}
\usepackage{epsfig}
\usepackage{graphicx}
\usepackage{amsmath}
\usepackage{amssymb}
\usepackage{soul}

\usepackage[pagebackref=true,breaklinks=true,letterpaper=true,colorlinks,bookmarks=false]{hyperref}

\cvprfinalcopy 

\def\cvprPaperID{7601} 

\ifcvprfinal\fi
\begin{document}

\title{RMP-SNN: Residual Membrane Potential Neuron for Enabling Deeper High-Accuracy and Low-Latency Spiking Neural Network}

\author{Bing Han, Gopalakrishnan Srinivasan, and Kaushik Roy\\
School of Electrical and Computer Engineering, Purdue University\\
{\tt\small \{han183,srinivg,kaushik\}@purdue.edu}
}

\maketitle

\begin{abstract}
   Spiking Neural Networks (SNNs) have recently attracted significant research interest as the third generation of artificial neural networks that can enable low-power event-driven data analytics. The best performing SNNs for image recognition tasks are obtained by converting a trained Analog Neural Network (ANN), consisting of Rectified Linear Units (ReLU), to SNN composed of integrate-and-fire neurons with ``proper" firing thresholds. The converted SNNs typically incur loss in accuracy compared to that provided by the original ANN and require sizable number of inference time-steps to achieve the best accuracy. We find that performance degradation in the converted SNN stems from using ``hard reset" spiking neuron that is driven to fixed reset potential once its membrane potential exceeds the firing threshold, leading to information loss during SNN inference. We propose ANN-SNN conversion using ``soft reset" spiking neuron model, referred to as Residual Membrane Potential (RMP) spiking neuron, which retains the ``residual" membrane potential above threshold at the firing instants. We demonstrate near loss-less ANN-SNN conversion using RMP neurons for VGG-16, ResNet-20, and ResNet-34 SNNs on challenging datasets including CIFAR-10 (93.63\% top-1), CIFAR-100 (70.93\% top-1), and ImageNet (73.09\% top-1 accuracy). Our results also show that RMP-SNN surpasses the best inference accuracy provided by the converted SNN with ``hard reset" spiking neurons using 2-8$\times$ fewer inference time-steps across network architectures and datasets.
\end{abstract}

\vspace{-0.05in}

\section{Introduction}
\label{sec:intro}

Deep neural networks, referred to as Analog Neural Networks (ANNs) in this article, composed of several layers of interconnected neurons, have achieved state-of-the-art performance in various Artificial Intelligence (AI) tasks including image localization and recognition \cite{krizhevsky2012imagenet, redmon2016you}, video analytics \cite{ngiam2011multimodal}, and natural language processing \cite{johnson2017google}, among other tasks. The superior performance has been achieved by trading off computational efficiency. For instance, ResNet \cite{kaiming2015resnet} that won the ImageNet Large Scale Visual Recognition Challenge in 2015 consists of 152 layers with over 60 million parameters, and incurs 11.3 billion FLOPS per classification. In an effort to explore more power efficient neural architectures, recent research efforts have been directed towards devising computing models inspired from biological neurons that compute and communicate using spikes. These emerging class of networks with increased bio-fidelity are known as Spiking Neural Networks (SNNs)\cite{maass1997networks}. The intrinsic power-efficiency of SNNs stems from their sparse spike-based computation and communication capability, which can be exploited to achieve higher computational efficiency in specialized neuromorphic hardware  \cite{blouw2019benchmarking, davies2018loihi, merolla2014million}.

\begin{figure}[!t]
\centering
\includegraphics[width=3.3in]{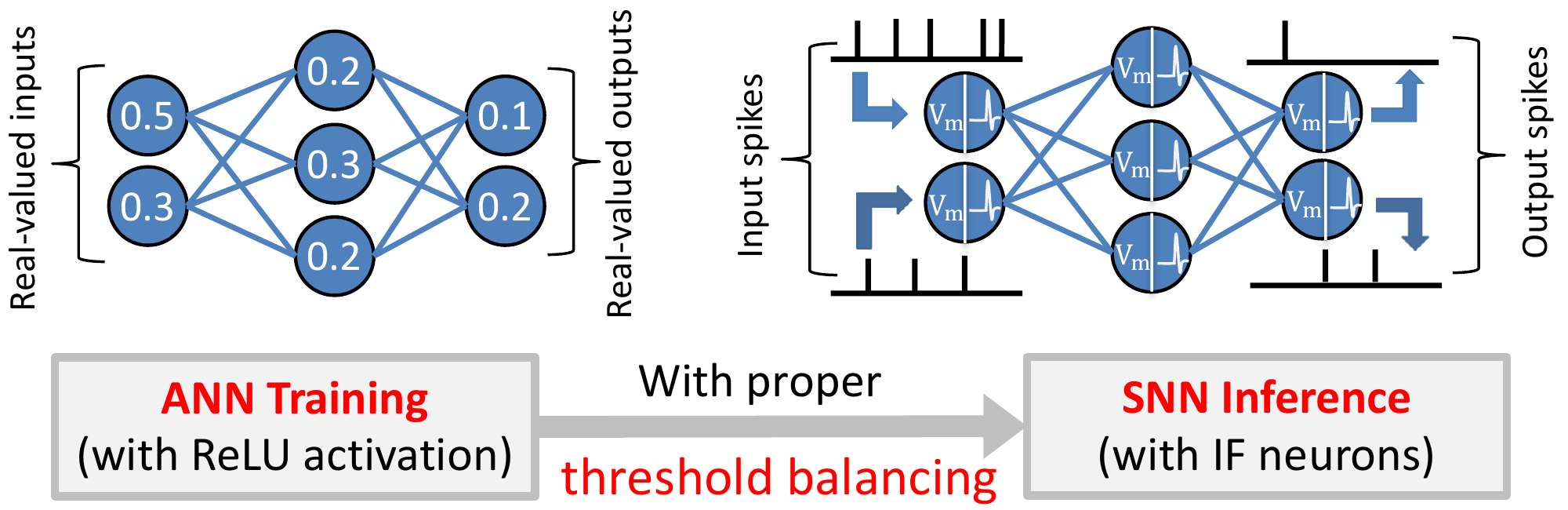}
\caption{Illustration of the ANN-SNN conversion methodology.}
\label{fig:ANN-SNN}
\end{figure}

Considering the rapid strides in accuracy achieved by ANNs over the past few years, SNN training algorithms are much less mature and are an active field of research. The training algorithms for SNNs can be categorized into Spike Timing Dependent Plasticity (STDP) based localized learning rules, spike-based error backpropagation, and ANN-SNN conversion methodologies. STDP-based unsupervised \cite{diehl2015unsupervised, masquelier2007unsupervised, srinivasan2018stdpbased, thiele2018event} and semi-supervised learning algorithms \cite{kheradpisheh2018stdp, lee2018deep, mozafari2018combining, tavanaei2018training} have thus far been restricted to shallow SNNs (with $\le$5 layers) yielding considerably lower accuracy than that provided by ANNs on complex datasets like CIFAR-10 \cite{ferre2018unsupervised, srinivasan2019restocnet}. In order to scale the networks much deeper, spike-based error backpropagation algorithms have been proposed for the supervised training of SNNs \cite{bellec2018long, jin2018hybrid, lee2019enabling, lee2016training, neftci2019surrogate, panda2016unsupervised, shrestha2018slayer, wu2018spatio}. The training complexity incurred for performing error backpropagation over time has limited their scalability for SNNs beyond 9-11 layers \cite{lee2019enabling}.

ANN-SNN conversion has yielded the best performing SNNs (typically composed of Integrate-and-Fire (IF) neurons), which are converted from a trained non-spiking ANN (consisting of Rectified Linear Unit (ReLU) as the activation function) \cite{cao2015spiking, diehl2015fast, diehl2016conversion, perez2013mapping, sengupta2019going, zhao2014feedforward} as illustrated in Fig. \ref{fig:ANN-SNN}. The conversion schemes intelligently assign ``appropriate" firing thresholds to the neurons at different layers of the network, thereby, ensuring that the IF spiking rates (number of spikes over large enough time interval) are proportional to the corresponding analog ReLU activations. Such conversion approaches take full advantage of backpropagation-based training, well-developed for ANNs. Note, however, the converted SNNs do not have the accuracy of the corresponding ANNs and require a sizeable number of time-steps ($>$2000 for ImageNet \cite{sengupta2019going}) for achieving the best accuracy (69.96\% \cite{sengupta2019going}). We find that performance degradation in the converted network stems from using spiking IF neurons, with ``hard reset" mechanism, to map analog activations to spike rates. A ``hard reset" neuron is driven to \textit{a priori} fixed low potential once its internal state (or membrane potential) exceeds the firing threshold, irrespective of how high the membrane potential is above the threshold. We find that ignoring the ``residual" potential above threshold leads to information loss in the conversion process (from ReLU-based artificial neurons to ``hard reset" IF neurons).

We propose conversion-based training using ``soft reset" spiking neuron, referred to as Residual Membrane Potential (RMP) spiking neuron, which better mimics the ReLU functionality. The RMP neuron keeps the ``residual" potential above firing threshold at the spiking instants instead of ``hard reset" to fixed potential, thereby alleviating the information loss that occurs during ANN-SNN conversion. We implemented deep SNN architectures such as VGG-16 and residual networks (ResNet-20 and ResNet-34) using RMP spiking neurons and demonstrate near loss-less conversion with close to state-of-the-art accuracy on complex datasets including ImageNet. We note that RMP neurons have been used for realizing deep SNNs that are converted from non-spiking ANNs \cite{rueckauer2016theory, rueckauer2017conversion}, albeit with higher conversion loss during SNN inference. We present the appropriate threshold initialization scheme to achieve near loss-less mapping of ReLU activations to RMP neuron spiking rates, yielding SNNs that provide the best inference accuracy to date on CIFAR-10, CIFAR-100, and ImageNet datasets. In addition, we demonstrate the ability of RMP-SNN to offer competitive accuracy using up to 8$\times$ fewer inference time-steps compared to converted SNN with ``hard reset" neurons, with only 1-2\% increase in overall spiking activity.

\section{Related Work}
\label{sec:related_work}

ANN-SNN conversion has been shown to be promising approach for building deep SNNs yielding high enough accuracy for complex image recognition \cite{cao2015spiking, diehl2015fast, perez2013mapping, rueckauer2016theory, rueckauer2017conversion, sengupta2019going, zhao2014feedforward} and natural language processing tasks \cite{diehl2016conversion}. The conversion schemes train ANN, composed of ReLU non-linearity, using backpropagation with added constraints like removal of bias neurons and batch normalization layers. The trained ANN is mapped to SNN composed of IF neurons. A notable exception to ReLU-IF mapping is the work of Hunsberger et al. \cite{hunsberger2015spiking} who used more bio-plausible Leaky-Integrate-and-Fire (LIF) neuron during inference by training the ANN with rate-based soft-LIF non-linearity. Efficient ANN-SNN conversion requires careful initialization of thresholds at every layer of the network so that the spiking rates are proportional to the ReLU activations. In this regard, Deihl et al. \cite{diehl2015fast} proposed model-based and data-based threshold balancing schemes. Model-based scheme estimates the threshold using only the ANN weights while the data-based scheme uses both the training data and weights. Following the work of \cite{diehl2015fast}, Sengupta et al. \cite{sengupta2019going} proposed data-based scheme that additionally uses SNN spiking statistics to achieve much improved ANN-SNN conversion, which has been shown to scale well to complex datasets like ImageNet. However, the aforementioned approaches are inherently susceptible to information loss during inference due to the use of ``hard reset" neurons as will be explained in section \ref{sec:ReLU_IF}. Rueckauer et al. \cite{rueckauer2017conversion} attempted to mitigate the information loss by using ``soft reset" RMP neurons. However, they report significantly high accuracy loss (14.28\%) between VGG-16 ANN and SNN on ImageNet compared to that (1.13\%) incurred by the approach of Sengupta et al. \cite{sengupta2019going} using ``hard reset" neurons. We believe that the higher accuracy loss incurred by Rueckauer et al. \cite{rueckauer2017conversion} is a consequence of unconstrained training of ANN with bias neurons and batch normalization layers as also pointed out in \cite{sengupta2019going}. While removing the constraints improved the accuracy of ANN, the converted SNN suffered from substantial accuracy loss, thereby, hiding the potential benefits of RMP neurons. We perform constrained ANN training and demonstrate near loss-less low-latency ANN-SNN conversion with ``soft reset" RMP neurons. The novelty of our work is the proposal of ANN-SNN conversion methodology using a combination of ``soft reset" RMP spiking neuron, appropriate layer-wise threshold initialization, and constrained ANN training (removal of batch normalization layers and bias neurons) to enable near loss-less ANN-SNN conversion.

\section{ANN-SNN Conversion}
\label{sec:conversion_preliminaries}

The fundamental distinction between ANN and SNN is the notion of time. In ANNs, input and output of neurons in all the layers are real-valued, and inference is performed with single feed-forward pass through the network. On the other hand, input and output of spiking neurons are encoded temporally using sparse spiking events over certain time period. Hence, inference in SNNs is carried out over multiple feed-forward passes or time-steps (also known as inference latency), where each pass entails sparse spike-based computations. Achieving close to ANN accuracy with minimal inference latency is key to obtaining favorable trade-off between accuracy and computational efficiency. The proposed conversion methodology significantly advances the state-of-the-art in this regard as will be detailed in section \ref{sec:results}.

\subsection{Input Encoding for SNNs}
\label{sec:input_encoding}

We use Poisson rate coding to map the input image pixels to spike trains firing at a rate (number of spikes over time) proportional to the corresponding pixel intensities as shown in \cite{heeger2000poisson}. The pixel intensity is first mapped to instantaneous spiking probability of the corresponding input neuron. We use Poisson process to generate the input spike in a stochastic manner as explained below. At every time-step of SNN operation, we generate a uniform random number between 0 and 1, which is compared against the neuronal firing probability. A spike is produced if the random number is less than the neuronal firing probability. Note that, input images fed to ANN are typically normalized to zero mean and unit standard deviation, yielding pixel intensities between $\pm 1$. For the SNN, we generate positive or negative spikes based on the sign of the normalized intensities. The time interval (inference latency) is dictated by the desired accuracy.



\begin{figure}[!t]
\centering
\includegraphics[width=2.6in]{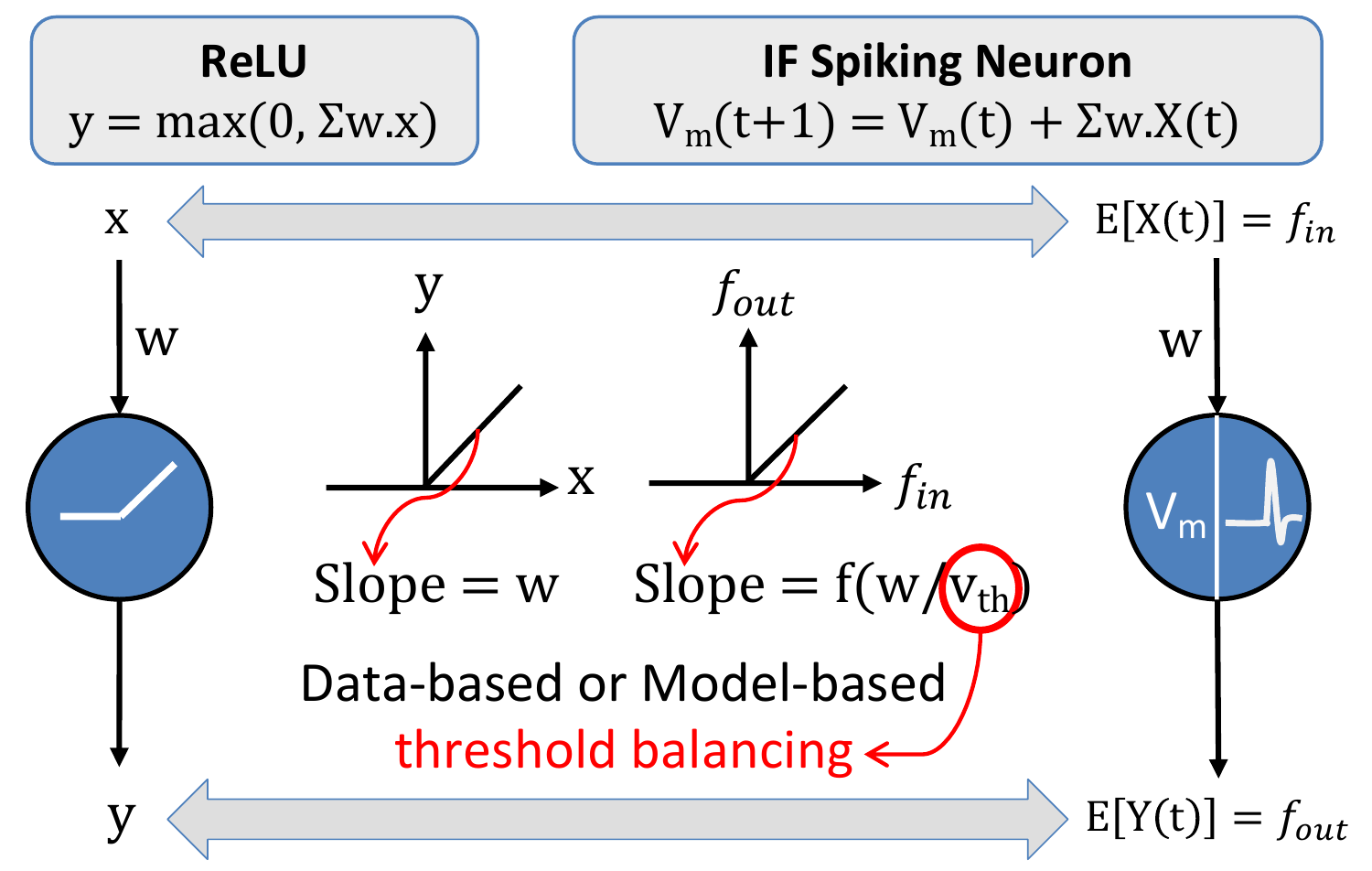}
\caption{Illustration of ReLU-IF mapping. The IF neuron threshold is set using model-based \cite{diehl2015fast} or data-based schemes \cite{diehl2015fast, sengupta2019going} so that its output rate is proportional to the ReLU activation.}
\label{fig:ReLU_IF}
\end{figure}

\subsection{ReLU-IF Mapping with Hard Reset}
\label{sec:ReLU_IF}

\begin{figure}[!t]
\centering
\includegraphics[width=3.3in]{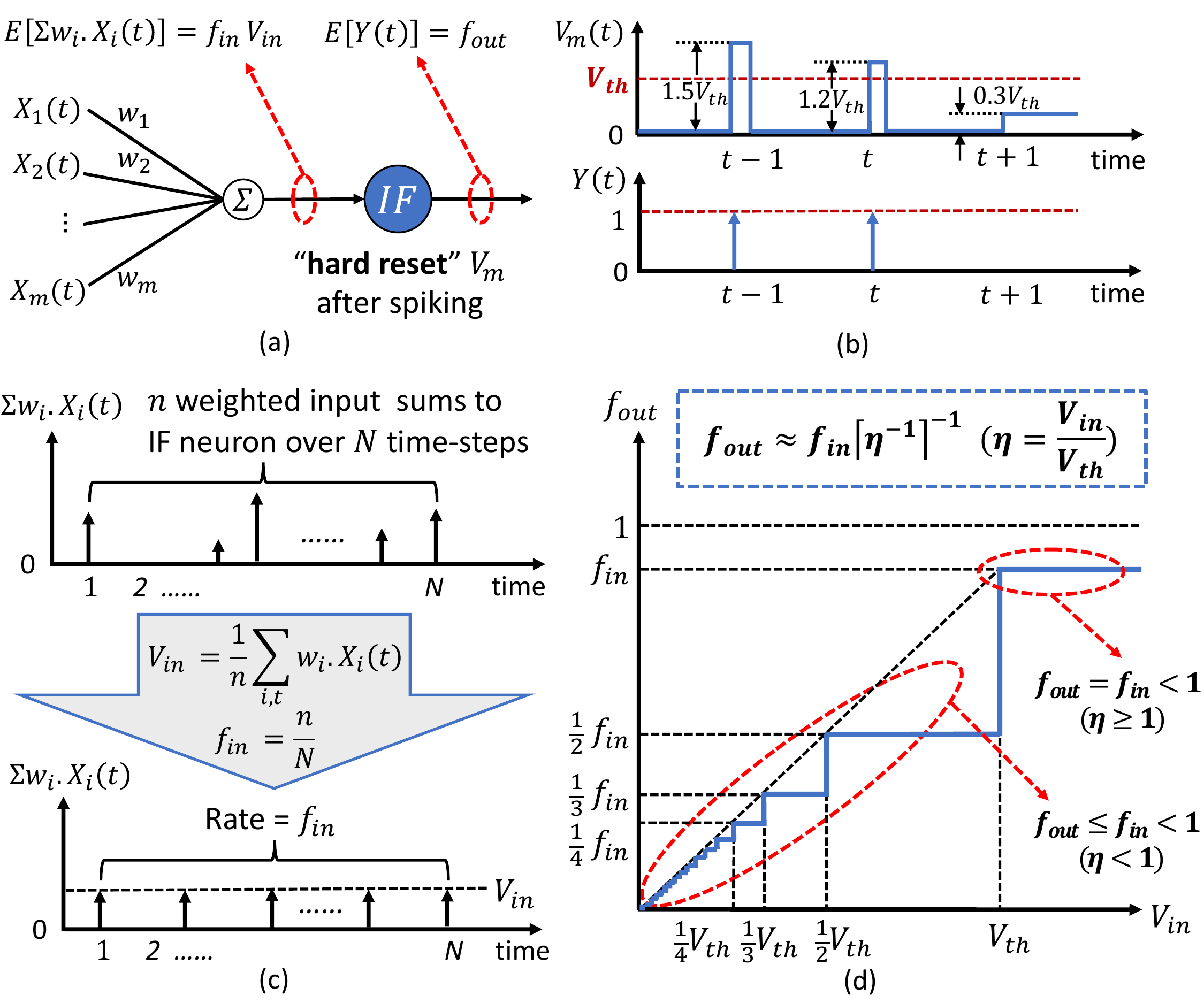}
\caption{(a) ``Hard reset" IF neuron driven by a set of input neurons via weights ($w$). (b) Illustration of reduced firing rate of IF neuron due to resetting the membrane potential ($V_m$) to zero at the spiking instants. (c) Mapping the expectation of weighted input sum received by the IF neuron over time to the product of average rate $f_{in}$ and amplitude $V_{in}$. (d) Non-linear input-output ($f_{in}$-$f_{out}$) response of IF neuron for different $V_{in}$.}
\label{fig:IF_dynamics}
\end{figure}

ANNs used for conversion to SNNs are typically trained with ReLU non-linearity \cite{nair2010rectified}, which is described by
\begin{equation}
\begin{aligned}
Y = max \left(0,X\right)
\end{aligned}
\end{equation}
where $Y$ is the output of ReLU-based artificial neruon, $X=\sum_i w_i.x_i+b$ is the weighted sum of input $x_i$ with weight $w_i$ and bias $b$. The bias is usually set to zero for effective ANN-SNN conversion \cite{sengupta2019going}. The ReLU ouput varies linearly with positive inputs. The linear ReLU dynamics are roughly mimicked using Integrate-and-Fire (IF) neuron as illustrated in Fig. \ref{fig:ReLU_IF}. An IF neuron receives train of spikes, over certain time period, whose rate corresponds to the real-valued ReLU input. The IF neuron integrates the weighted spike-input into its membrane potential whose dynamics are described by
\begin{equation}
    V_m(t) = V_m(t-1) + \sum_i w_i.X_i(t)
\end{equation}
where $V_m(t)$ is the membrane potential at time-step $t$, $w_i$ is the transferred weight from ANN, and $X_i$ is the spike train of $i$-th input neuron. The IF neuron produces a spike when its membrane potential exceeds the firing threshold $V_{th}$ ($>$ 0), which is estimated using model-based \cite{diehl2015fast} or data-based schemes \cite{diehl2015fast, sengupta2019going}. At the instant of a spike, the membrane potential is ``hard reset" to $0$ irrespective of the amount by which the membrane potential exceeds the threshold. Ignoring the residual membrane potential above threshold affects the expected linear relationship between the input and the output spiking rates as illustrated below with an example. Let us suppose that an IF neuron (shown in Fig. \ref{fig:IF_dynamics}(a)) receives weighted input sum of 1.5$V_{th}$, 1.2$V_{th}$, and 0.3$V_{th}$ in three successive time-steps as depicted in Fig. \ref{fig:IF_dynamics}(b). The total weighted input sum across the three time-steps is 3$V_{th}$. The IF neuron needs to fire thrice to maintain precise linear relationship between the input and output spiking rates. In effect, it generates only two spikes over three time-steps as a result of ignoring the residual potential above threshold at the firing instants as illustrated in Fig. \ref{fig:IF_dynamics}(b).

We now formalize the deviation of the input-output relationship of ``hard reset" neuron from the expected linear behavior. The average weighted input sum, $E[\sum_i w_i.X_i(t)]$, received by the IF neuron can be specified as $f_{in}\,V_{in}$, where $f_{in}$ and $V_{in}$ are the mean rate and amplitude of the weighted input sum, respectively, as shown in Fig. \ref{fig:IF_dynamics}(c). The average input amplitude, $V_{in}$, can moreover be specified as $\eta\,V_{th}$ for $\eta \in \mathbb{R^+}$ without any loss of generality. The output firing rate (number of output spikes over time), $f_{out}$, of the IF neuron is then described by

\begin{equation}
\ f_{out} = \begin{cases}
        f_{in} \indent \indent \ \ \ \ \ \ \ & \eta = \frac{V_{in}}{V_{th}} \geq 1 \\
        \lfloor f_{in} \lceil \eta^{-1} \rceil^{-1} N\rfloor N^{-1}  & 0 \leq \eta < 1 \\
        \approx f_{in} \lceil \eta^{-1} \rceil^{-1} & 0 \leq \eta < 1 \ and \ N \gg 1 
    \end{cases}\\
\label{eq:fout_IF}
\end{equation}

in  which  $\lceil \ \rceil$  is the ceiling operation, $\lfloor \ \rfloor$ is the floor operation, $f_{out} \leq f_{in} \leq 1$, and $N$ is the total number inference time-steps. The output rate matches the input rate only for $\eta \ge 1$ when the average input amplitude $V_{in}$ is larger than the threshold $V_{th}$. In this case, the input amplitude is high enough to warrant an output spike every time it occurs in spite of ignoring the residual potential from earlier spiking instants. On the other hand, when $0 \leq \eta < 1$, the output spiking rate $f_{out}$ is approximately specified by $\lceil \eta^{-1} \rceil^{-1} f_{in}$ (for large enough $N$) as described in equation \ref{eq:fout_IF}. The ceiling operation accounts for the non-linear relationship between the input and output rates as illustrated in Fig. \ref{fig:IF_dynamics}(d) and explained below. Consider $\eta \in [\frac{1}{k+1}, \frac{1}{k})$ and $V_{in} \in [\frac{V_{th}}{k+1}, \frac{V_{th}}{k})$ for any positive integer $k$. As the average input amplitude, $V_{in}$, gradually changes from $\frac{V_{th}}{k+1}$ to $\frac{V_{th}}{k}$, the output rate $f_{out}$ remains constant at $\frac{f_{in}}{k+1}$ instead of linearly increasing to $\frac{f_{in}}{k}$ as shown in Fig. \ref{fig:IF_dynamics}(d). For instance, let us suppose that $\eta \in [\frac{1}{2}, 1)$ and $V_{in} \in [\frac{V_{th}}{2}, V_{th})$ while $f_{in}$ is unity. If $V_{in}$ of $\frac{V_{th}}{2}$ is received per time-step, the IF neuron fires a spike every second time-step. When $V_{in}$ increases to $\frac{3\,V_{th}}{4}$ per time-step, the IF neuron still fires a spike only every second time-step instead of firing $3$ spikes over $4$ time-steps. The reduced output rate is a direct consequence of ignoring the residual potential at the firing instants by performing ``hard reset" to $0$.

Fig. \ref{fig:IF_dynamics}(d) shows that roughly linear input-output relationship can be obtained for $\eta \ll 1$, which requires the average input amplitude to be much lower than the firing threshold of ``hard reset" IF neurons. Low input activity in converted SNN can be achieved by setting the layer-wise thresholds to be much higher, which reduces the ANN-SNN conversion loss at the cost of significantly high inference latency. On the other hand, lowering the thresholds, to minimize the inference latency, increases the layer-wise input spiking activity, resulting in $\eta$ closer to $1$. Higher $\eta$ causes the ``hard reset" neurons to operate in the non-linear regime, leading to larger degradation in SNN accuracy. ANN-SNN conversion with ``hard reset" neurons requires careful initialization of thresholds to obtain favorable accuracy-latency trade-off.

\section{Residual Membrane Potential SNN}
\label{sec:RMP_SNN}

\begin{figure}[!t]
\centering
\includegraphics[width=3.3in]{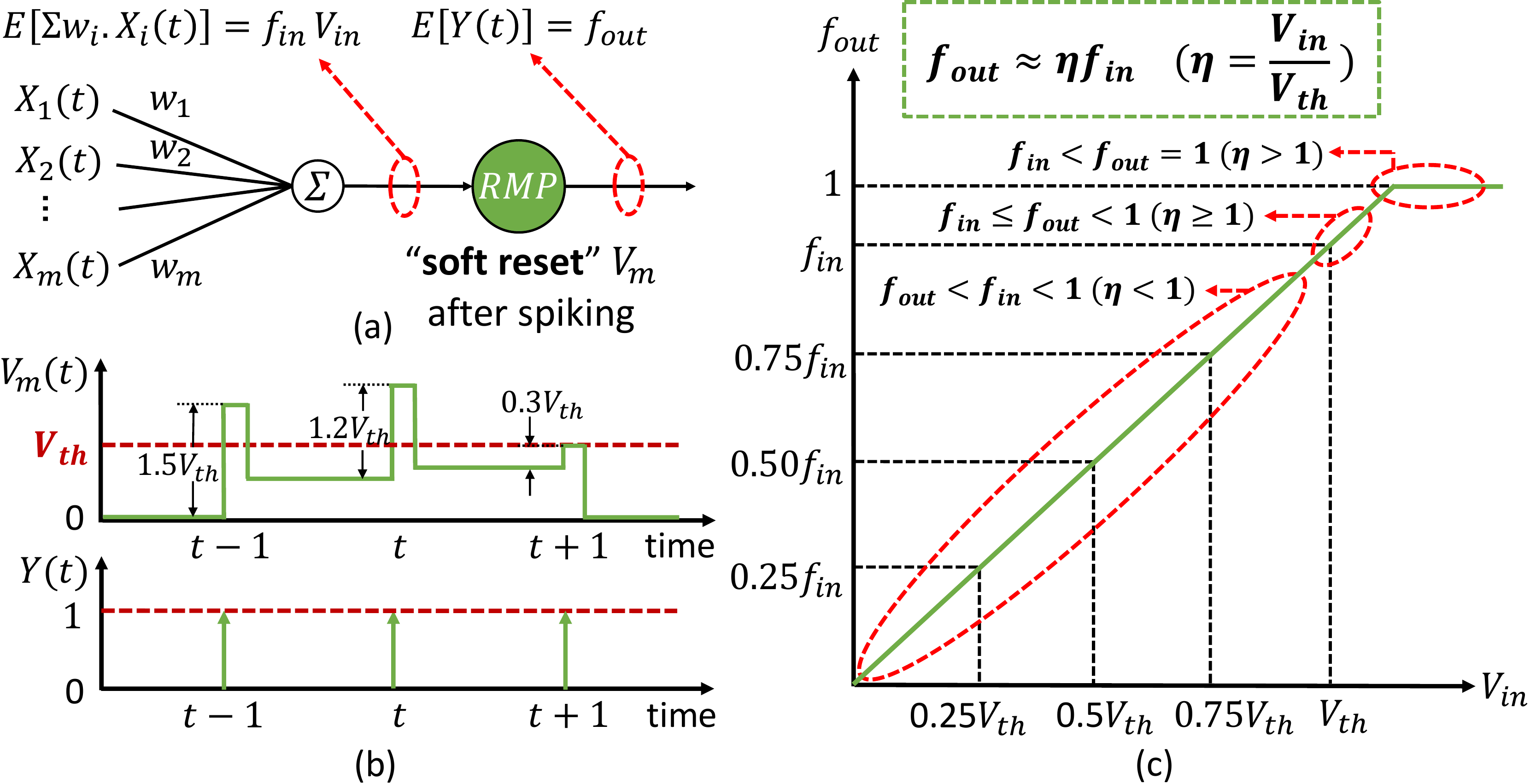}
\caption{(a) ``Soft reset" RMP neuron driven by a set of input neurons via weights ($w$). (b) Illustration of precise spiking behavior of RMP neuron by retaining the residual potential at the firing instants. (c) Linear input-output ($f_{in}$-$f_{out}$) response of RMP neuron for different $V_{in}$.}
\label{fig:RMP_dynamics}
\end{figure}

We propose Residual Membrane Potential (RMP) spiking neuron (shown in Fig. \ref{fig:RMP_dynamics}(a)) to obtain linear input-output characteristics and achieve near loss-less ANN-SNN conversion. The RMP neuron minimizes information loss during inference by performing ``soft reset" as described in the following pseudo-code.

\begin{tabular}{l l}
          &  \\
       1: & if \ $V_{m}(t) \geq V_{th}:$  \\
       2: & \indent Emit\ Output\ Spike: $Y(t)=1$ \\
       3: & \indent Perform\ Soft\ Reset: $V_m(t)=V_m(t)-V_{th}$ \\
          &  \\
\end{tabular}

At the instant of a spike, the membrane potential ($V_m$) is reduced by an amount equal to the firing threshold ($V_{th}$) instead of ``hard reset" to $0$. ``Soft reset" effectively retains the residual potential above threshold as shown in Fig. \ref{fig:RMP_dynamics}(b). Let us suppose that an RMP neuron receives weighted input sum of 1.5$V_{th}$, 1.2$V_{th}$, and 0.3$V_{th}$, totalling to 3$V_{th}$, across three consecutive time-steps. It produces the expected number of three spikes by retaining the residual potential at the firing instants as depicted in Fig. \ref{fig:RMP_dynamics}(b). Note that ``soft reset" is also referred to as ``reset by subtraction" in SNN literature \cite{rueckauer2016theory, rueckauer2017conversion}. Formally, the output firing rate, $f_{out}$, of RMP neuron can be described by
\begin{equation}
\ f_{out} = \begin{cases}
\lfloor \eta f_{in} N \rfloor N^{-1} &\ \eta \geq 0 \\
\approx \eta f_{in} &\ \eta \geq 0 \ and \ N \gg 1 
\end{cases}\\
\label{eq:fout_RMP}
\end{equation}

in which $f_{in} \leq 1, \ f_{out} \leq 1$, $\eta = \frac{V_{in}}{V_{th}}$ is the ratio between the average amplitude of weighted input sum $V_{in}$ and firing threshold $V_{th}$, and $N$ is the inference latency. The output rate changes proportional to the input rate by a factor $\eta$ for a wide range of $\eta$ as depicted in Fig. \ref{fig:RMP_dynamics}(c) due to carrying over the residual potential at the firing instant to the following time-step. The linear input-output characteristics exhibited by the RMP neuron for a wide range of $\eta$ enable it to provide near loss-less ANN-SNN mapping for a wide range of firing thresholds as will be discussed in the following section \ref{sec:threshold_init}.


\subsection{Threshold Balancing for RMP-SNN}
\label{sec:threshold_init}

ANN-SNN conversion requires assigning ``appropriate" threshold for the spiking neurons to ensure that they operate in the linear (or almost linear) regime, which effectively leads to lower (or even negligible) conversion loss. The extended linear input-output relationship of RMP neuron (see Fig. \ref{fig:RMP_dynamics}(c)) provides ``wider operating range" for the neuronal firing threshold compared to that for the ``hard reset" IF neuron (refer to Fig. \ref{fig:IF_dynamics}(d)). This begets the following couple of questions that need to be answered to ensure appropriate threshold balancing for the RMP neuron.
\begin{enumerate}
    \item For any given $f_{in}$ and $V_{in}$, what is the desired operating range for the RMP neuron firing threshold to ensure loss-less ANN-SNN conversion?
    \item How should the absolute value of threshold be determined so that the RMP neuron operates in the desired range?
\end{enumerate}
We determine the upper and lower bounds for the RMP neuron firing threshold based on the desired operating range for the output rate $f_{out}$. Fig. \ref{fig:RMP_dynamics}(c) indicates that the RMP neuron allows $f_{out}$ to be higher than the input rate $f_{in}$, while still exhibiting linear input-output dynamics. The desirable range for $f_{out}$ is $[f_{in}, 1)$. This is because $f_{out}{\ge}f_{in}$ ensures sufficient spiking activity across successive layers of deep SNN, leading to high enough accuracy using fewer inference time-steps. Satisfying $f_{out}{\ge}f_{in}$ requires $\eta{=}\frac{V_{in}}{V_{th}}\ge 1$ or $V_{th}{\le}V_{in}$ as highlighted in Fig. \ref{fig:RMP_dynamics}(c). On the other hand, $f_{out}{<}f_{in}$ (or $V_{th}{>}V_{in}$) leads to gradual reduction in SNN spiking activity with network depth, thereby, increasing the inference latency.

Next, the lower bound for $V_{th}$ is determined to ensure that the output rate $f_{out}$ is less than unity. This is because,  $f_{out}{\ge}1$ produces a spike at every time-step irrespective the received input. The excessive spiking activity can lead to substantial degradation in accuracy. The threshold required for guaranteeing $f_{out}{<}1$ can be obtained from the following equation that relates the average input and output potentials, which is described by
\begin{equation}
    \frac{dV_{m}^{avg}}{dt} = f_{in} V_{in}-f_{out} V_{th}
\label{eq:Vm_reservior}
\end{equation}
where $V_m^{avg}$ is the average membrane potential of the RMP neuron. In order for the average potential to reach its steady state value, $\frac{dV_{m}^{avg}}{dt}$ in \eqref{eq:Vm_reservior} must be equal to zero. The output rate $f_{out}$ is then specified by the following equation.
\begin{equation}
    f_{out} = \frac{f_{in} V_{in}}{V_{th}}
\label{eq:output_rate}
\end{equation}
Equation \ref{eq:output_rate} clearly indicates that $V_{th}$ must be greater than $f_{in} V_{in}$ for $f_{out}$ to be smaller than $1$. Thus, the desired operating range of $V_{th}$, for given $f_{in}$ and $V_{in}$, is specified by
\begin{equation}
    f_{in}V_{in} \leq V_{th} \leq V_{in}
\label{eq:threshold_balancing}
\end{equation}
which answers the first question raised in the beginning of this section. As explained previously, the $V_{th}$ range specified in \eqref{eq:threshold_balancing} ensures $f_{in}{\le}f_{out}{<}1$, which can lead to optimal accuracy-latency trade-off. It is important to note that $f_{out}$ for ``hard reset" IF neuron is always smaller than or equal to input rate $f_{in}$ (as shown in Fig. \ref{fig:IF_dynamics}(d)) due to ignoring the residual potential. As a result, the ``hard reset" IF neuron inherently incurs higher inference latency compared to that for the RMP neuron.

\begin{figure}[!t]
\centering
\includegraphics[width=3.3in]{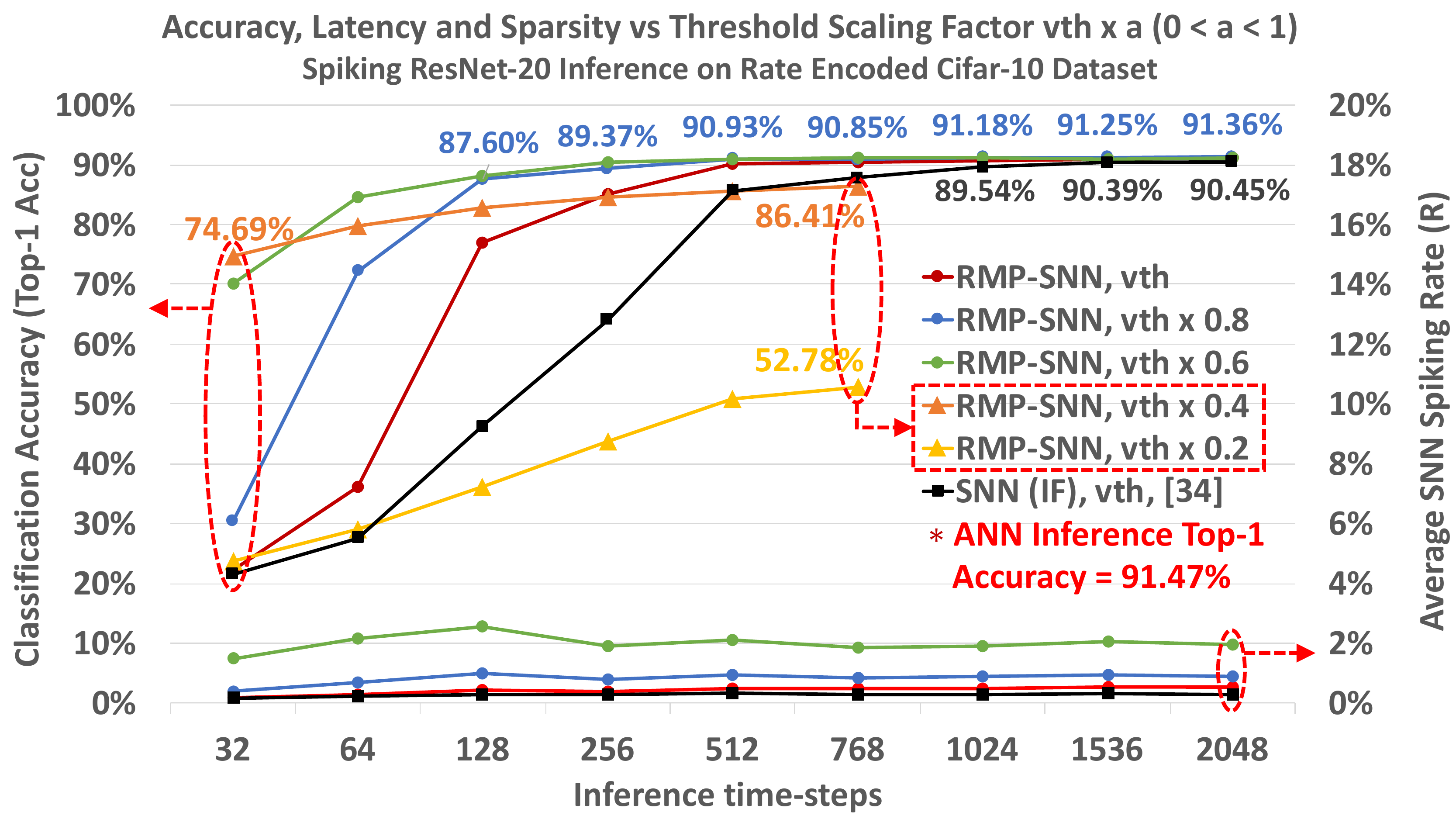}
\caption{Inference accuracy and spiking activity versus latency of ResNet-20 SNN, composed of RMP neurons, for different threshold scaling factor $\alpha$ on the CIFAR-10 dataset.}
\label{fig:resnet20_cifar10}
\end{figure}

We now address the second question concerning the precise $V_{th}$ estimation methodology. In our analysis thus far, we estimated $V_{th}$ using $V_{in}{=}E[\sum_i w_{i}X_{i}(t)]$, which is the average weighted input sum to the RMP neuron over time. Prior works proposed setting $V_{th}$ to $Max_t[\sum_i w_{i}X_{i}(t)]$, which is the maximum weighted input sum to the neuron across time-steps \cite{diehl2015fast, sengupta2019going}. The maximum estimate $V_{in}^{max}$ can enable the RMP neuron to operate in the linear region (where $f_{out}{=}f_{in}$ as highlighted in Fig. \ref{fig:RMP_dynamics}(c)), while the average estimate $V_{in}^{avg}$ ($< V_{in}^{max}$) can cause it to operate more in the vicinity of the non-linear region (where $f_{out}{=}1$ designated in Fig. \ref{fig:RMP_dynamics}(c)). We validate our hypothesis of using $V_{in}^{max}$ versus $V_{in}^{avg} (\approx \alpha V_{in}^{max}$ where $\alpha \in (0,1)$ is a scaling factor) using ResNet-20 SNN on the CIFAR-10 dataset. Before presenting the results, we describe the methodology, originally proposed in \cite{sengupta2019going}, used to initialize the layer-wise threshold of deep SNN using the ANN-trained weights and SNN spiking statistics. We transfer the trained weights from ANN to SNN, and feed the Poisson spike-inputs (for the entire training set) to the first layer of the SNN. We record the weighted input sum to all the neurons in the first layer across time-steps. We set the threshold of RMP neurons in the first layer to the maximum weighted input sum, across neurons and time-steps, over the training dataset. We then freeze the threshold of the first layer, and estimate the threshold of the second layer using the same procedure outlined previously. The threshold estimation process is carried out sequentially in a layer-wise manner for all the layers.

ResNet-20 SNN, with its layer-wise threshold assigned to $V_{in}^{max}$, achieved 91.36\% on CIFAR-10, which is comparable to that (91.47\%) achieved by the corresponding ANN as illustrated in Fig. \ref{fig:resnet20_cifar10}. We thereafter scaled the threshold by a factor of up to 0.6$\times$ and found that the RMP-SNN, with scaled threshold, converged to the same accuracy obtained using $V_{th}{=}V_{in}^{max}$. This corroborates our hypothesis that the RMP neuron operates in the linear region for a wide range of firing thresholds, thereby, causing the RMP-SNN to yield higher accuracy using fewer time-steps as depicted in Fig. \ref{fig:resnet20_cifar10}. As the threshold is scaled further by up to 0.2$\times$, we notice significant drop in accuracy. At such low thresholds, the RMP neuron operates in the non-linear (excessive spiking) regime, leading to higher accuracy loss during inference. We propose initializing the threshold of RMP-SNN with scaled version of $V_{in}^{max}$ (scaling factor $\alpha \le$ 0.6 in this example) to achieve the optimal accuracy-latency trade-off. We validate the presented threshold initialization scheme across different SNN architectures and datasets.

Improving the inference latency by reducing the firing threshold increases the spiking activity, thereby, adversely impacting the overall computational efficiency. In an effort to quantify the spiking activity of RMP-SNN for different thresholds, we measure the average spike rate as defined by the following equation.
\mathchardef\mhyphen="2D
\begin{equation}
    R = \frac{total \ spikes}{total \ neurons \times \ inference \ \mathit{time \mhyphen steps}} \times 100\%
\label{eq:spiking_rate}
\end{equation}
The spike rate $R$ in \eqref{eq:spiking_rate} indicates the average percentage of neurons that spike per time-step. Our analysis indicates that the RMP-SNN, with scaled thresholds, provides disproportionate benefits in accuracy and latency compared to the increase in spiking activity ($\sim$1-2\%) as will be discussed in section \ref{sec:results}.

\section{Results}
\label{sec:results}
We evaluated RMP-SNNs on standard visual object recognition benchmarks, namely the CIFAR-10, CIFAR-100 and ImageNet datasets. We use VGG-16 architecture \cite{Simonyan15} for all three datasets. ResNet-20 configuration outlined in \cite{kaiming2015resnet} is used for the CIFAR-10 and CIFAR-100 datasets while ResNet-34 is used for experiments on the ImageNet dataset. Our implementation is derived from the Facebook ResNet implementation code for CIFAR and ImageNet datasets. The code can be found online at \url{https://github.com/facebookarchive/fb.resnet.torch}. Proper weight initialization is crucial to achieve convergence in such deep networks without batch-normalization. Similar weights initialization was done as outlined in \cite{HardtM16} although their networks were trained without both dropout and batch-normalization. For VGG networks, a dropout layer is used after every ReLU layer except for those layers which are followed by a pooling layer. For Residual networks, we use dropout only for the ReLUs at the non-identity parallel paths but not at the junction layers. We found this to be crucial for achieving training convergence. 

\begin{figure}[!b]
\centering
\includegraphics[width=3.3 in]{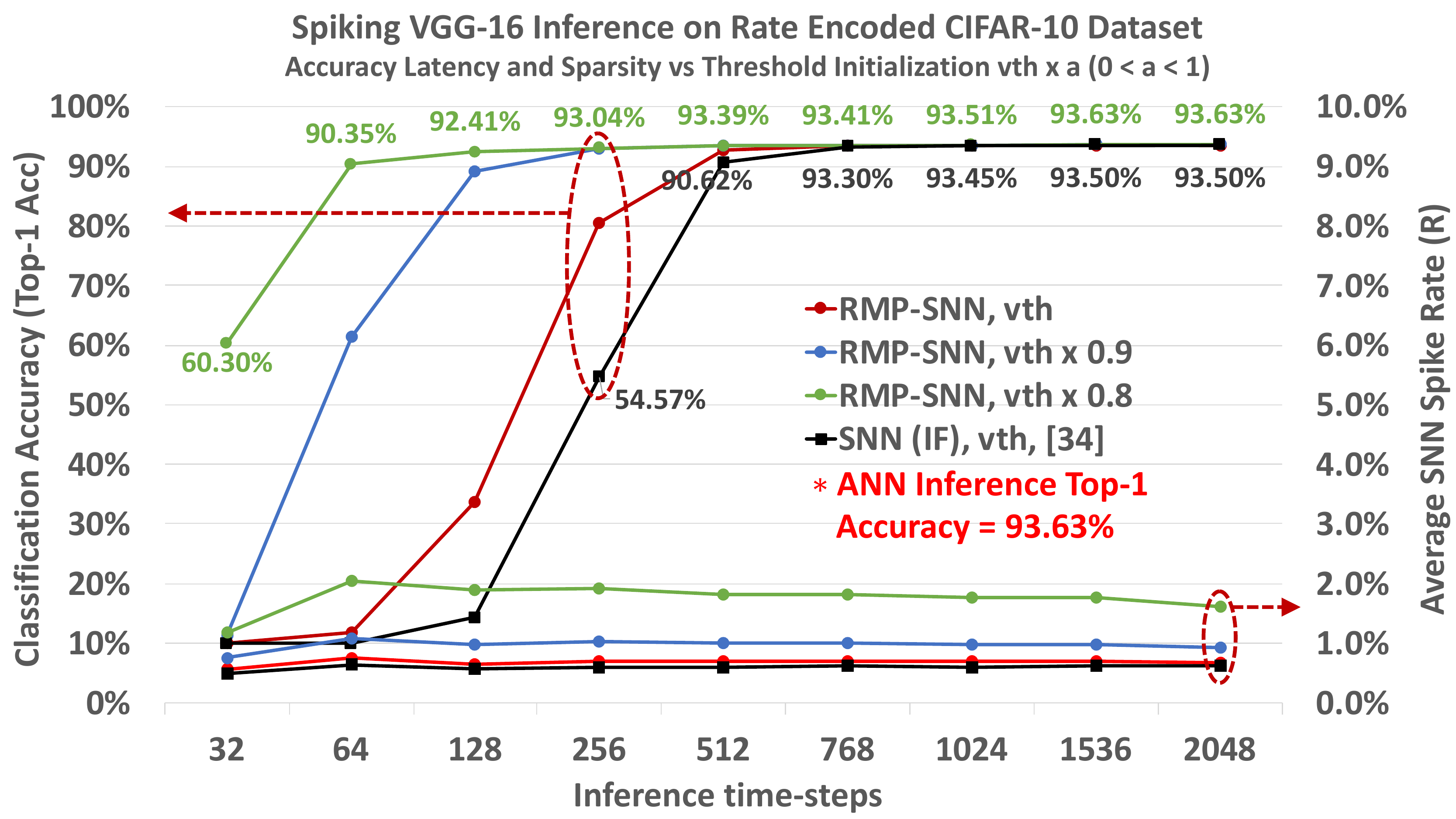}
\caption{Inference performance comparison between the VGG-16 RMP-SNN and the baseline VGG-16 SNN (IF) on CIFAR-10 dataset.}
\label{vgg16_cifar10_Ab}
\end{figure}

\begin{figure}[!b]
\centering
\includegraphics[width=3.3 in]{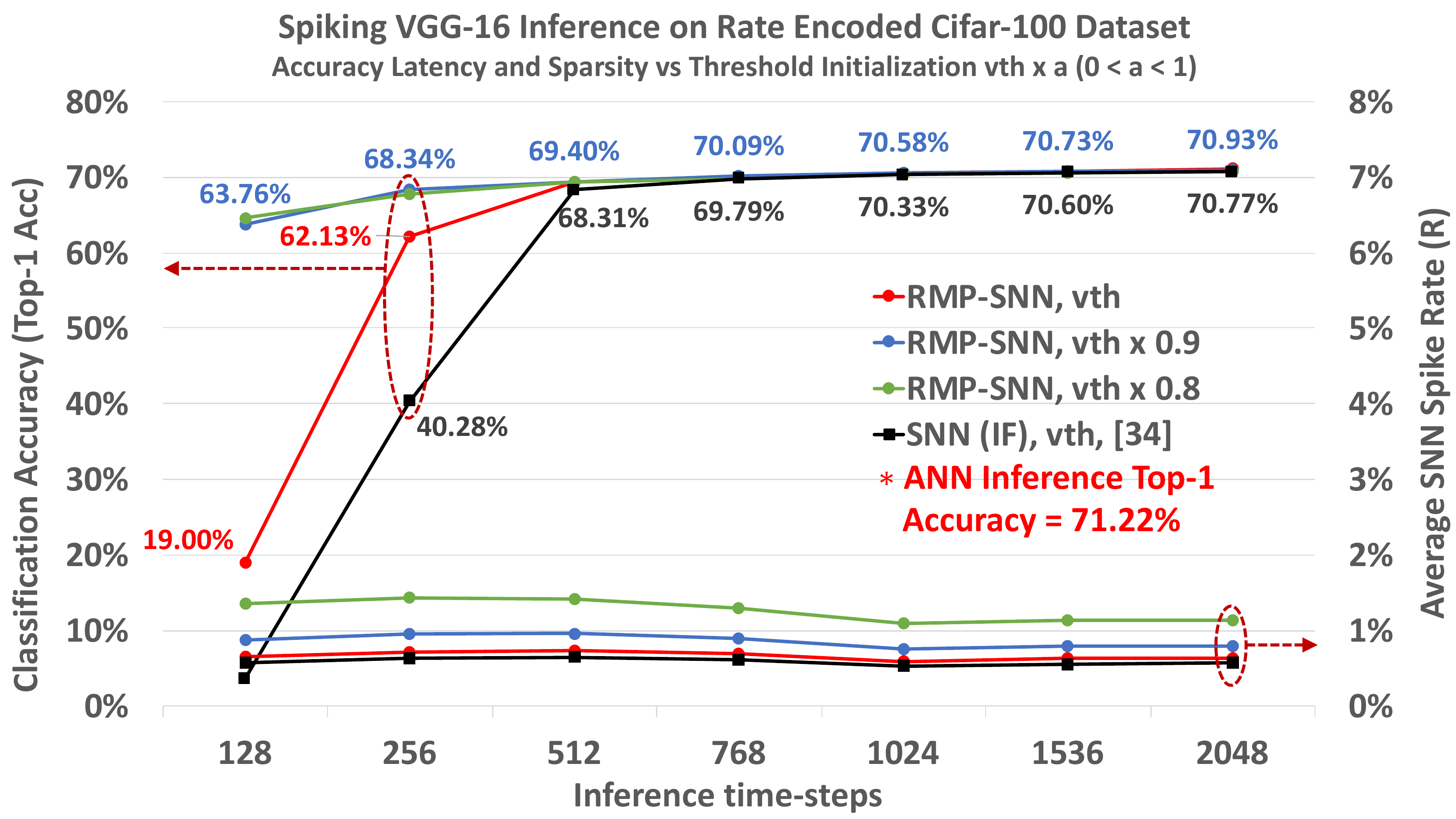}
\caption{Inference performance comparison between the VGG-16 RMP-SNN and the baseline VGG-16 SNN (IF) on CIFAR-100 dataset.}
\label{vgg16_cifar100}
\end{figure}

\begin{figure}[!t]
\centering
\includegraphics[width=3.3in]{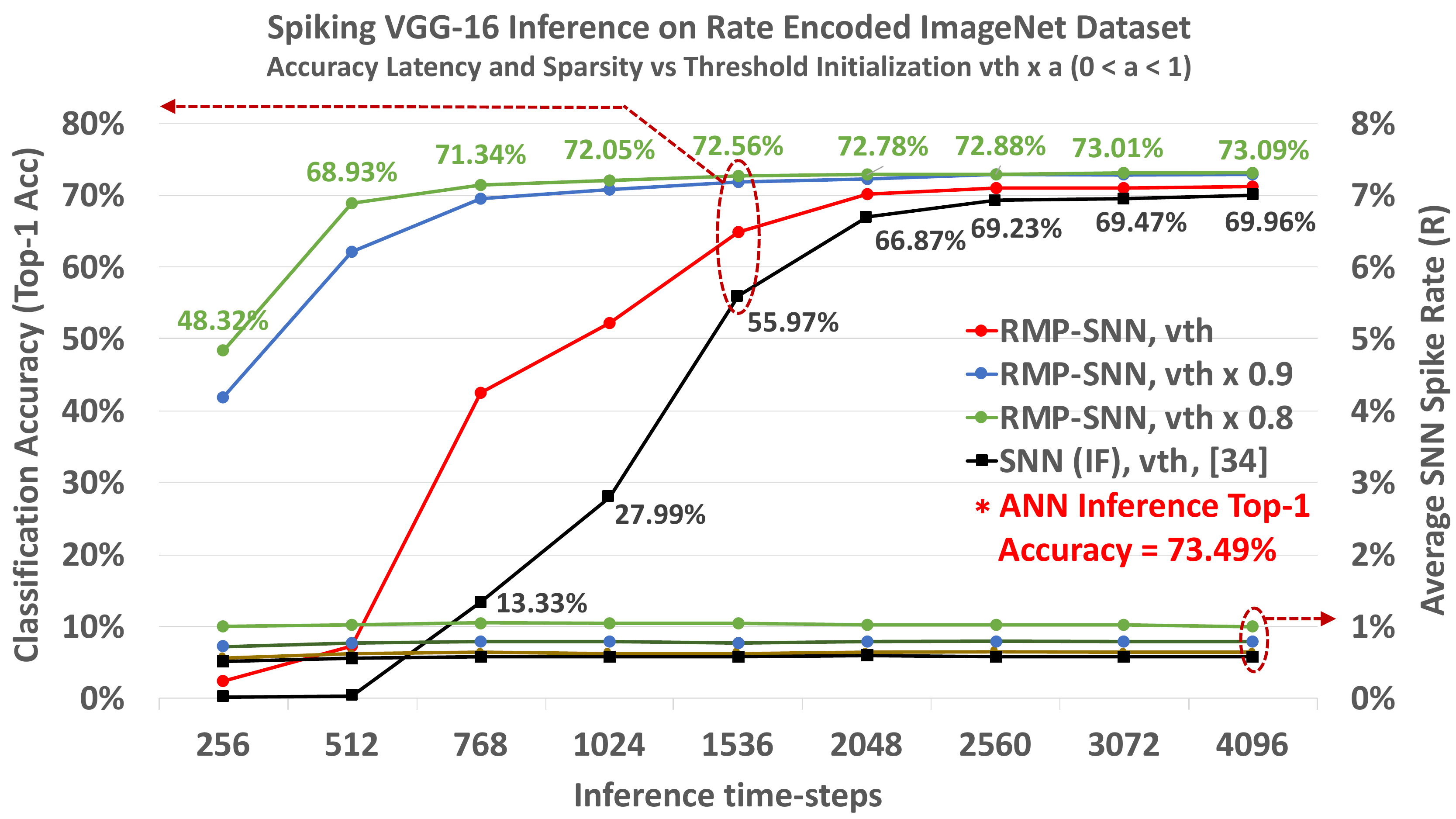}
\caption{Inference performance comparison between the VGG-16 RMP-SNN and the baseline VGG-16 SNN (IF) on ImageNet dataset.}
\label{vgg16_imagenet}
\end{figure}

\begin{figure}[!t]
\centering
\includegraphics[width=3.3in]{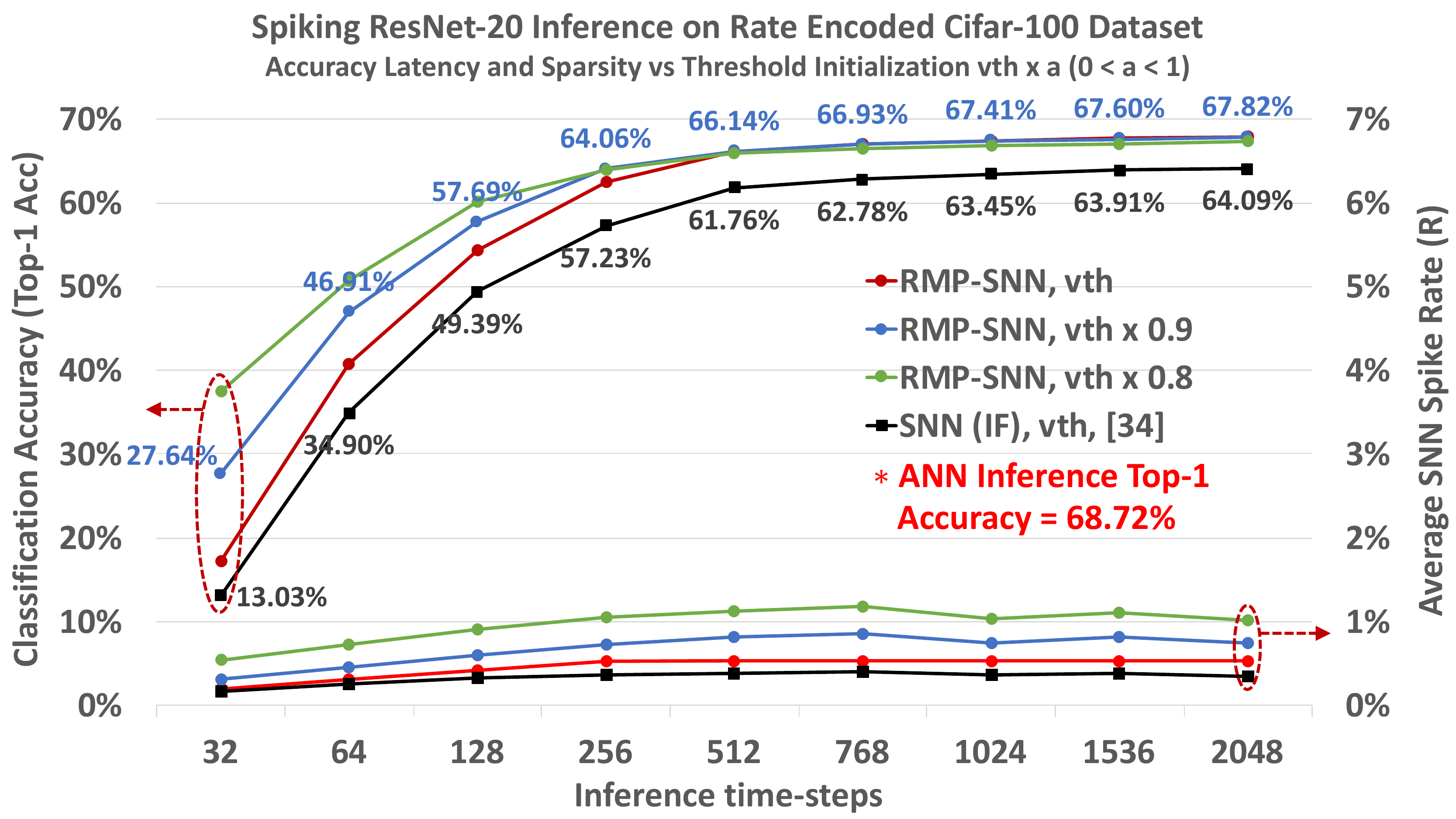}
\caption{Inference performance comparison between the ResNet-20 RMP-SNN and the baseline ResNet-20 SNN (IF) on CIFAR-100 dataset.}
\label{resnet20_cifar100}
\end{figure}

\begin{figure}[!t]
\centering
\includegraphics[width=3.3in]{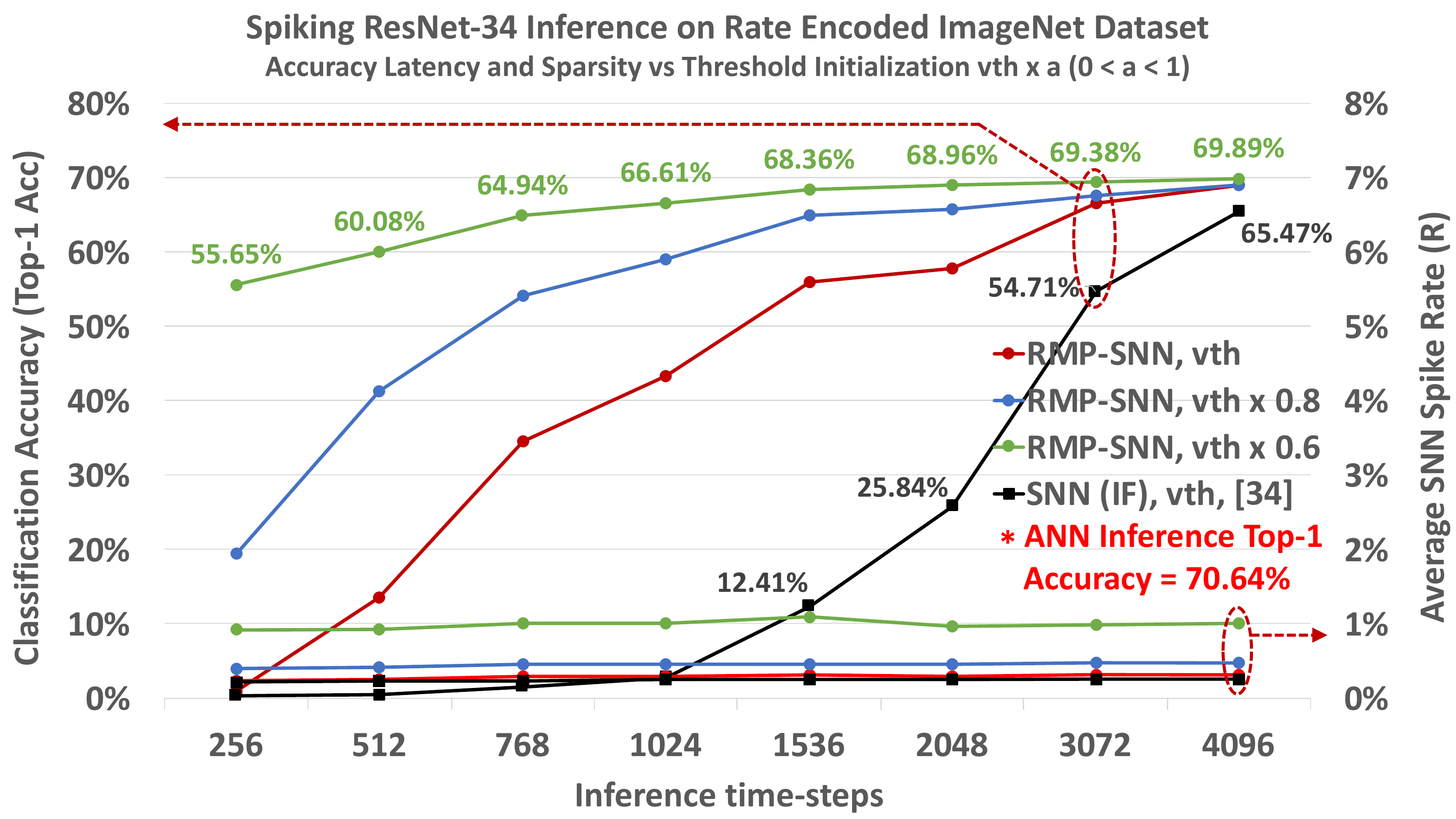}
\caption{Inference performance comparison between the ResNet-34 RMP-SNN and the baseline ResNet-34 SNN (IF) on ImageNet dataset.}
\label{resnet34_imagenet}
\end{figure}

The most recent state-of-the-art ANN-SNN conversion works are provided for comparison as shown in Table.\ref{tb:SNNs-accuracy-cifar10}, \ref{tb:SNNs-accuracy-cifar100} and \ref{tb:SNNs-accuracy-imagenet}. Note that authors in \cite{rueckauer2017conversion} reported a top-1 SNN error rate of 25.04\% for an Inception-V3 network, with their ANN trained to an error rate of 23.88\%. The resulting conversion loss is 1.52\% which is much higher than our proposal. The Inception-V3 network conversion was also optimised by a voltage clamping method, that was found to be specific for the Inception network and did not apply to the VGG network \cite{rueckauer2017conversion}. In addition, the results reported on ImageNet in \cite{rueckauer2017conversion} are on a subset of 1382 image samples for Inception-V3 network and 2570 samples for VGG-16 network. Hence, the performance on the entire dataset is unclear. Our proposed RMP-SNN achieved not only the best SNN inference accuracies but also the lowest ANN-SNN conversion loss across all network architectures and datasets we evaluated. All SNN results reported represent the average of 5 independent runs. RMP-SNNs performances on accuracy, latency and sparsity are also presented and compared with the best performing SNNs to date in \cite{sengupta2019going} as shown in Fig.\ref{fig:resnet20_cifar10} to Fig.\ref{resnet34_imagenet}. In each figure, x-axis is the SNN inference latency, the y-axis on the left measures the SNN top-1 inference accuracy, and the y-axis on the right measures the average spike rate.

\begin{table*}[!t]
\centering
\caption{Accuracy loss due to ANN-SNN conversion of the state-of-the-art SNNs on CIFAR-10 dataset}
\label{tb:SNNs-accuracy-cifar10}
\resizebox{0.8\textwidth}{!}{%
\begin{tabular}{|l|l|c|c|c|}
\hline
  \textbf{Network Architecture} & \textbf{Spiking Neuron Model} & \textbf{ANN (Top-1 Acc)} & \multicolumn{1}{l|}{\textbf{SNN (Top-1 Acc)}} & \multicolumn{1}{l|}{\textbf{Accuracy Loss}}\\ \hline
  8-layered \cite{HunsbergerE16} & LIF (hard-reset) & 83.72\% & 83.54\% & 0.18\% \\ \hline
  3-layered \cite{EsserMACAABMMBN16} & LIF (hard-reset) & - & 89.32\% & - \\ \hline
  6-layered \cite{rueckauer2017conversion} & IF (hard-reset) & 91.91\% & 90.85\% & 1.06\% \\ \hline
  ResNet-20 \cite{sengupta2019going} & IF (hard-reset) & 89.1\% & 87.46\% & 1.64\% \\ \hline
  \textbf{ResNet-20} \textbf{[This Work]} & \textbf{RMP (soft-reset)} & \textbf{91.47\%} & \textbf{91.36\%} & \textbf{0.11\%} \\ \hline
  VGG-16 \cite{sengupta2019going} & IF (hard-reset) & 91.7\% & 91.55\% & 0.15\% \\ \hline
  \textbf{VGG-16} \textbf{[This Work]} & \textbf{RMP (soft-reset)} & \textbf{93.63\%} & \textbf{93.63\%} & $\boldsymbol{<}$ \textbf{0.01\%} \\ \hline
\end{tabular}%
}
\end{table*}

\begin{table*}[!t]
\centering
\caption{Accuracy loss due to ANN-SNN conversion of the state-of-the-art SNNs on CIFAR-100 dataset}
\label{tb:SNNs-accuracy-cifar100}
\resizebox{0.8\textwidth}{!}{%
\begin{tabular}{|l|l|c|c|c|}
\hline
  \textbf{Network Architecture} & \textbf{Spiking Neuron Model} & \textbf{ANN (Top-1 Acc)} & \multicolumn{1}{l|}{\textbf{SNN (Top-1 Acc)}} & \multicolumn{1}{l|}{\textbf{Accuracy Loss}}\\ \hline
  ResNet-20 \cite{sengupta2019going} & IF (hard-reset) & 68.72\% & 64.09\% & 4.63\% \\ \hline
  \textbf{ResNet-20} \textbf{[This Work]} & \textbf{RMP (soft-reset)} & \textbf{68.72\%} & \textbf{67.82\%} & \textbf{0.9\%} \\ \hline
  VGG-16 \cite{sengupta2019going} & IF (hard-reset) & 71.22\% & 70.77\% & 0.45\% \\ \hline
  \textbf{VGG-16} \textbf{[This Work]} & \textbf{RMP (soft-reset)} & \textbf{71.22\%} & \textbf{70.93\%} & \textbf{0.29\%} \\ \hline
\end{tabular}%
}
\end{table*}

\begin{table*}[!t]
\centering
\caption{Accuracy loss due to ANN-SNN conversion of the state-of-the-art SNNs on ImageNet dataset}
\label{tb:SNNs-accuracy-imagenet}
\resizebox{0.8\textwidth}{!}{%
\begin{tabular}{|l|l|c|c|c|}
\hline
 \textbf{Network Architecture} & \textbf{Spiking Neuron Model} & \textbf{ANN (Top-1 Acc)} & \multicolumn{1}{l|}{\textbf{SNN (Top-1 Acc)}} & \multicolumn{1}{l|}{\textbf{Accuracy Loss}}\\ \hline
 ResNet-34 \cite{sengupta2019going} & IF (hard-reset) & 70.69\% & 65.47\% & 5.22\% \\ \hline
 \textbf{ResNet-34} \textbf{[This Work]} & \textbf{RMP (soft-reset)} & \textbf{70.64\%} & \textbf{69.89\%} & \textbf{0.75\%} \\ \hline
 VGG-16 \cite{rueckauer2017conversion} & RMP (soft-reset)  & 63.89\% & 49.61\% & 14.28\% \\ \hline
 VGG-16 \cite{sengupta2019going} & IF (hard-reset) & 70.52\% & 69.96\% & 0.56\% \\ \hline
 \textbf{VGG-16} \textbf{[This Work]} & \textbf{RMP (soft-reset)} & \textbf{73.49\%} & \textbf{73.09\%} & \textbf{0.4\%} \\ \hline
\end{tabular}%
}
\end{table*}

The VGG-16 RMP-SNN inference on CIFAR-10 dataset is shown in Fig.\ref{vgg16_cifar10_Ab}. RMP-SNN achieved the same accuracy 93.63\% as the trained ANN using 2048 time-steps, whereas the SNN with IF neurons achieved 93.50\% at the end of 2048 time-steps. The fastest RMP-SNN with reduced threshold (green curve) reaches an accuracy above 90\% using only 64 time-steps, which is 8 times faster than the baseline SNN with IF neurons that uses 512 time-steps. Reducing thresholds causes an increase in spike rate. However, the fastest RMP-SNN with reduced threshold (green curve) still attains a spike rate less than 2\%. Note, in this work, we reported higher accuracy of the baseline SNN with IF neurons compared to \cite{sengupta2019going}, in which, their best accuracy of the VGG-16 SNN with IF neurons on CIFAR-10 dataset is 91.55\%. This is because we trained the ANN to a higher accuracy than the one used in \cite{sengupta2019going} and the baseline SNN with IF neurons in our work is converted from the better trained ANN. 

The VGG-16 RMP-SNN inference on CIFAR-100 dataset is shown in Fig.\ref{vgg16_cifar100}, which reaches an accuracy of 70.93\% using 2048 time-steps, whereas the baseline SNN with IF neurons reaches 70.77\% at the end of 2048 time-steps. Note, no VGG-16 SNN was evaluated on CIFAR-100 dataset in \cite{sengupta2019going}. In this work, both RMP-SNN and the baseline SNN with IF neurons were converted from our trained ANN with top-1 inference accuracy of 71.22\%. The RMP-SNN with reduced threshold (blue curve) reaches an accuracy of 68.34\% using only 256 time-steps, which is 2 times faster than the baseline SNN with IF neurons that uses about 512 time-steps. The RMP-SNN with reduced threshold (blue curve) attains a spike rate lower than 1\% throughout the entire inference time-steps. 

The VGG-16 RMP-SNN inference on the ImageNet dataset is shown in Fig.\ref{vgg16_imagenet}. It reaches an accuracy of 73.09\% using 4096 time-steps, whereas the SNN with IF neurons reaches 69.96\% using 4096 time-steps. Both RMP-SNN and the baseline SNN with IF neurons are converted from our trained ANN with top-1 inference accuracy of 73.49\%. The RMP-SNN with reduced threshold (green curve) reaches an accuracy of 68.93\% using only 512 time-steps, which is 4.5 times faster than the baseline SNN with IF neurons using over 2300 time-steps. The RMP-SNN with reduced threshold (green curve) attains a spike rate as low as 1\% throughout the entire inference time-steps.

The ResNet-20 ANN has been trained to have top-1 inference accuracy of 91.47\% on CIFAR-10 dataset as shown in Fig.\ref{fig:resnet20_cifar10} (in section \ref{sec:threshold_init}). After conversion, the RMP-SNN reaches top-1 accuracy of 91.36\% using 2048 time-steps, whereas the SNN with IF neurons reaches 90.45\% using the same 2048 time-steps. The RMP-SNN with reduced threshold (green curve) reaches an accuracy above 85\% using only 64 time-steps, which is 8 times faster than the baseline SNN with IF neurons, that uses 512 time-steps. The RMP-SNN with reduced threshold (green curve) attains a spike rate around 2\% throughout the inference time-steps.

The trained ResNet-20 ANN has top-1 inference accuracy of 68.72\% on CIFAR-100 dataset as shown in Fig.\ref{resnet20_cifar100}. The RMP-SNN reaches top-1 accuracy of 67.82\% using 2048 time-steps, whereas the SNN with IF neurons reaches top-1 accuracy 64.09\% using the same 2048 time-steps. The fastest RMP-SNN with reduced threshold (green curve) reaches an accuracy of 64.06\% using only 256 time-steps, which is 8 times faster than the baseline SNN with IF neurons that uses about 2048 time-steps. The fastest RMP-SNN with reduced threshold (green curve) attains a spike rate about 1\% throughout the inference time-steps.

The trained ResNet-34 ANN has top-1 inference accuracy of 70.64\% on the ImageNet dataset as shown in Fig.\ref{resnet34_imagenet}. The RMP-SNN reaches an accuracy of 69.89\% using 4096 time-steps, whereas the SNN with IF neurons reaches 65.47\% using the same 4096 time-steps. The fastest RMP-SNN with reduced threshold (green curve) reaches an accuracy of 60.08\% using only 512 time-steps, which is 7 times faster than the baseline SNN with IF neurons that uses more than 3500 time-steps. The fastest RMP-SNN with reduced threshold (green curve) attains a spike rate as low as 1\% throughout the inference time-steps.

\section{Conclusion and Discussion}
In this work, we propose an ANN to SNN conversion technique. It uses novel spiking neuron model named RMP spiking neuron that retains a residual membrane potential after firing. The RMP spiking neuron better mimics the ReLU functionality than the IF neuron by allowing a residual potential to remain after the neuron has fired, alleviating the information loss that occurs during the ReLU to IF conversion. We also propose a threshold balancing technique which alleviates the spike rate vanishing issue in SNNs and significantly improved the latency and scalability of RMP-SNNs to very deep architectures. We implemented large scale deep network architectures such as VGG and Residual networks using the proposed conversion based training and evaluated performance on cifar-10, cifar-100 and ImageNet datasets. Our proposed RMP-SNNs achieve the best accuracies and lowest conversion loss than the state-of-the-art across all network architectures and datasets we tested. 

\section*{Acknowledgment}
This work was supported in part by C-BRIC, Center for Spintronic Materials, Interfaces, and Novel Architectures (C-SPIN), a MARCO and DARPA sponsored StarNet center, by the Semiconductor Research Corporation, National Science Foundation, Sandia National Laboratories, Vannevar Bush Faculty Fellowship and by the US Army Research Laboratory and the UK Ministry of Defense under Agreement Number W911NF-16-3-0001.

{\small
\bibliographystyle{ieee_fullname}
\bibliography{egpaper_for_review}
}

\end{document}